\documentclass[final]{cvpr}

\usepackage{graphics} 
\usepackage{epsfig} 
\usepackage{mathptmx} 
\usepackage{times} 
\usepackage{amsmath} 
\usepackage{amssymb}  
\usepackage{xcolor}
\usepackage{etoolbox}
\usepackage{url}    
\usepackage{fancyhdr}   
\usepackage{ifthen}
\usepackage{textcomp}
\usepackage{gensymb} 
\usepackage{makecell}
\usepackage{stmaryrd} 
\usepackage{graphicx}
\usepackage{cuted}
\usepackage{capt-of}
\usepackage{multirow}
\usepackage{multicol}
\usepackage{bigstrut}
\usepackage{adjustbox}
\usepackage{booktabs}
\usepackage{tabularx}
\usepackage[normalem]{ulem}
\usepackage[linesnumbered,ruled,vlined]{algorithm2e}
\usepackage{soul} 
\usepackage[colorlinks,citecolor=red,urlcolor=blue,bookmarks=false,hypertexnames=true]{hyperref} 
\DeclareMathOperator*{\argmin}{argmin}

\newcommand{\myfootnote}[2]{{%
  \let\thempfn\relax
  \footnotetext[0]{$^#1$\emph{#2}}
}}

\definecolor{red}{rgb}{0.9,0.1,0}
\definecolor{gray}{rgb}{0.8,0.8,0.8}
\definecolor{blue}{rgb}{0.4,0.4,0.9}
\definecolor{slateblue}{rgb}{0.7,0.35,0.9}
\definecolor{green}{rgb}{0, 0.4, 0}
\definecolor{orange}{rgb}{1, 0.5, 0}
\definecolor{mahogany}{rgb}{0.75, 0.25, 0.0}
\definecolor{purple}{rgb}{0.6, 0, 0.6}
\definecolor{darkgreen}{rgb}{0, 0.4, 0}
\definecolor{frenchblue}{rgb}{0.0, 0.45, 0.73}
\definecolor{goldenrod}{rgb}{0.85, 0.65, 0.13}
\newboolean{revising}
\setboolean{revising}{false}
\ifthenelse{\boolean{revising}}
{

    \newcommand{\ignore}[1]{}
    \newcommand{\kike}[1]{\textcolor{blue}{#1}}
    \newcommand{\kikereplace}[2]{\textcolor{gray}{\sout{#1}} {\textcolor{blue}{[kike]: #2}}}

} {

    \newcommand{\ignore}[1]{}
    \newcommand{\kike}[1]{{#1}}
    \newcommand{\kikereplace}[2]{#2}

}

\newcommand{\R}[1]{\mathbb{R}}

\def\cvprPaperID{800} 
\def\confYear{CVPR 2021}

\begin{document}

\title{\LARGE \bf
Robust 360-8PA: Redesigning The Normalized 8-point Algorithm for 360-FoV Images
}

\author{
    Bolivar Solarte$^{1}$ \\
    {\tt\small enrique.solarte.pardo@gamil.com}
    \and
    Chin-Hsuan Wu$^{1}$\\
    {\tt\small chinhsuanwu@gapp.nthu.edu.tw}
    \and
    Kuan-Wei Lu$^{1}$\\
    {\tt\small kuanweilu@gapp.nthu.edu.tw}
    \and
    Min Sun$^{1}$\\
    {\tt\small sunmin@ee.nthu.edu.tw}
    \and
    Wei-Chen Chiu$^{2}$\\
    {\tt\small walon@cs.nctu.edu.tw}
    \and
    Yi-Hsuan Tsai$^{3}$\\
    {\tt\small ytsai@nec-labs.com} 
}

\maketitle

\myfootnote{1}{National Tsing Hua University}
\myfootnote{2}{National Chiao Tung University }
\myfootnote{3}{NEC Labs America}


 \begin{abstract}

In this paper, we present a novel preconditioning strategy for the classic 8-point algorithm (8-PA)
for estimating an essential matrix from 360-FoV images (i.e., equirectangular images) in spherical projection.
To alleviate the effect of uneven key-feature distributions and \kike{outlier correspondences}, which can potentially \kikereplace{degenerate}{decrease} the accuracy of an essential matrix, our method optimizes a non-rigid transformation to deform a spherical camera into a new spatial domain, defining a new constraint and a more robust and accurate solution for an essential matrix.
\kikereplace{We provide extensive experiments to show that our method reduces camera pose errors under different levels of outlier correspondence and scene conditions, showing an accuracy increment up to 20\% compared to the 8-PA, without the significant overhead in computation time.}
{Through several experiments using random synthetic points, 360-FoV, and fish-eye images, we demonstrate that our normalization can increase the camera pose accuracy about 20\% without significantly overhead the computation time.} 
\kikereplace{In addition, we extend the usage of our method to spherical cameras by proposing both a constant weighted least-square optimization, which further improves over the well-known Gold Standard Method (i.e., the non-linear optimization of the camera pose recovered from an essential matrix), showing that our algorithm is a more reliable, robust, and accurate solution.}
{In addition, we present further benefits of our method through both a constant weighted least-square optimization that improves further the well known Gold Standard Method (GSM) (i.e., the non-linear optimization by using epipolar errors); and a relaxation of the number of RANSAC iterations, both showing that our normalization outcomes a more reliable, robust, and accurate solution.}
\end{abstract}
\section{Introduction}\label{sec:sec_intro_test}

\kike{Estimating the relative pose between different views of the same scene has been studied for decades in Computer Vision and Robotics. For instances, Visual Odometry (VO), Simultaneous Localization and Mapping (SLAM), Structure from Motion (SFM), among others, generally leverage this primary estimation for initializing the first camera poses, triangulate landmarks in 3D, re-localize the camera pose, and prune out  outliers correspondences from the system.
}

\kike{In practice, the goal of estimating a camera pose between two images relies on finding a geometry constraint for their pixels. This is known as the essential matrix or epipolar constraint. In general, the procedure of calculating an essential matrix includes two main steps: 1) An abundant number of salient pixels, i.e., \textit{key-features}, are extracted from each image, followed by matching them across different views; 2) Based on that correspondences, the essential matrix is calculated satisfying the epipolar constraints. Finally, after the essential matrix is derived, the relative camera pose can be recovered by singular value decomposition~\cite{scaramuzza2011tutoI, hartleyBoock2003}.}

%
\begin{figure}
\centering
\includegraphics[width=0.85\columnwidth]{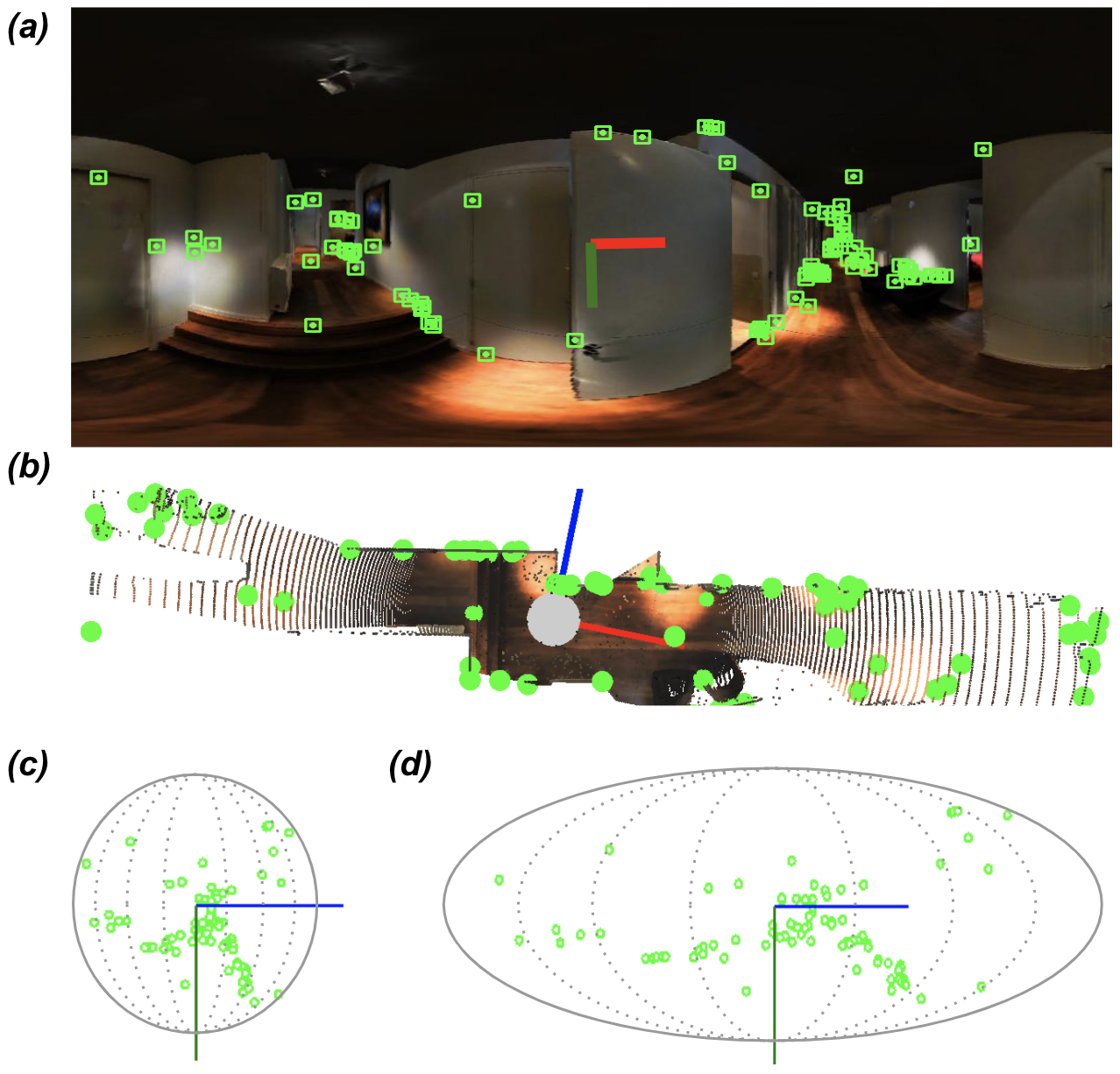}
\caption{\textbf{Illustration for the distribution of key-features in a 360-FoV image.} (a) and (b) present a 360-FoV image and its top view (point cloud) respectively, with its key-features denoted as landmarks in green color. In (c), the same landmarks presented in (a) and (b) are projected into a unit sphere. In (d), our proposed normalization scheme (cf. Section~\ref{sec:method:normalization}) has been applied to the spherical features in (c). Note that the key-features in the ovoid surface (d) are geometrically shifted compared to (c) due to our normalization. As a result, (d) expands the relative angles between every key-feature, which in turn leads to a more stable solution when using the 8-PA method~\cite{higgins1981computer}.}

\label{fig:teaser}
\end{figure}

\kike{Several algorithms have been proposed to find an essential matrix, however the most widely used solutions are the five-point (5-PA)~\cite{nister2004efficient} and eight-point (8-PA)~\cite{ higgins1981computer} algorithms. Despite the former uses the minimum number of correspondence points needed for calibrated cameras, its implementation usually relies on a polynomial approximation with multiple solutions. In contrast, the 8-PA is a linear method without the ambiguity in its outcome. In general, the 8-PA is mostly used for 360-FoV images (e.g., \cite{GuanSfM_VMF,openmvg_moulon2016, openvslam2019, fermuller1998spherical_SFM,fujiki2007epipolar,taira2015robust_features}), due to its simplicity and proved stability under large field of views \cite{da2019perturbation}. On the other hand, unlike the 5-PA, the 8-PA requires more iterations for outlier removal using a RANSAC evaluation, hence increasing its computation time for a large ratio of outliers \cite{scaramuzza2012tutoII}. This defines a clear disadvantage of the 8-PA.}


Although studies in~\cite{da2019perturbation} show that a wider FoV may increase the stability of the 8-PA for spherical cameras, these assume that matched key-points in the image are well-distributed in the whole FoV. However, in practice, that distribution mainly depends on external factors, which sometimes yields to clustered or uneven distribution of key-points (see Fig. \ref{fig:teaser})~(e.g. \cite{taira2015robust_features}).

In this work, we improve the 8-PA~\cite{higgins1981computer} for spherical cameras by re-defining \textit{the pre-conditioning} strategy proposed by~\cite{hartley1997defense} to be applied to spherical projection, which effectively deals with outliers and uneven key-feature distributions for 360-FoV images.
\kike{Additionally}, we extend the usage of our novel pre-conditioning by proposing a constant weighted least-square solution, which improves the Gold Standard Method (GSM)~\cite{pagani2011_sfm_some_loss_func, hartleyBoock2003} that is usually used to refine the camera pose. \kike{Lastly, we also present results comparing our solution under a RANSAC evaluation, showing that our preconditioning is capable to effectively deal with outliers, hence potentially reducing the number of required iterations.}

We evaluate our methods under both sequences of real 360-FoV and fish-eye scenarios, where the former is our own dataset, collected from Matterport3D~\cite{Matterport3D} and rendered using MINOS~\cite{minos2017}, while the latter is the TUM-VI dataset~\cite{TUMVIschubert2018dataset}. We show that our method significantly outperforms the state-of-the-art 8-PA for spherical cameras without the overhead in computation time, demonstrating the robustness of our method against noise and outliers.
In favor of the research community, the source code is available at~\textbf{\url{https://github.com/EnriqueSolarte/robust_360_8PA}}.


\section{Related Work} \label{sec:sec_relat}

Several approaches have been developed to estimate an essential matrix based on matched key-features. Nevertheless, the most well-known approaches are still the 5-PA~\cite{nister2004efficient} and 8-PA~\cite{higgins1981computer}, where the former has been improved largely in past years, e.g.,~\cite{5PA_Li_Hartley,li2006five,Li2013_4p_alg,Fathian2018Quest}. However, all of them rely on a polynomial approximation, which inevitably leads to multiple essential matrix solutions. In contrast, the 8-PA~\cite{higgins1981computer} is a simpler and linear method that always outputs a unique result.

In~\cite{hartley1997defense}, a normalization strategy is introduced upon 8-PA~\cite{higgins1981computer}, improving homogeneity in the input data, and robustness against noise. However, this preconditioning was mainly designed for uncalibrated pinhole cameras. Later, in~\cite{muhlich1998role}, the normalization~\cite{hartley1997defense} is further explained in terms of a generalized total least square problem, which opens the idea of a general normalization. However, in this work, the authors focus only on perspective cameras, \kike{proving that the normalization ~\cite{hartley1997defense}~(i.e., an isotropic and non-isotropic normalization), indeed, reduces the effects of noise in the solution of the 8-PA for key-point described in a homogeneous plane}. In the literature there is no evidence for normalization for spherical projection.

For spherical projection, the 8-PA has been widely used as a standard solution~(e.g.,~\cite{book_ma2012invitation, fermuller1998spherical_SFM, fujiki2007epipolar,pagani2011_sfm_some_loss_func,torii2005sphericalConcepts, GuanSfM_VMF}), that estimates a quick initial guess, where other methods can efficiently refine it further~\cite{scaramuzza2011tutoI, hartleyBoock2003, book_ma2012invitation}. Note that a reliable and fast estimation of an initial guess is always desired.  

\begin{figure*}
\centering
\includegraphics[width=1\linewidth]{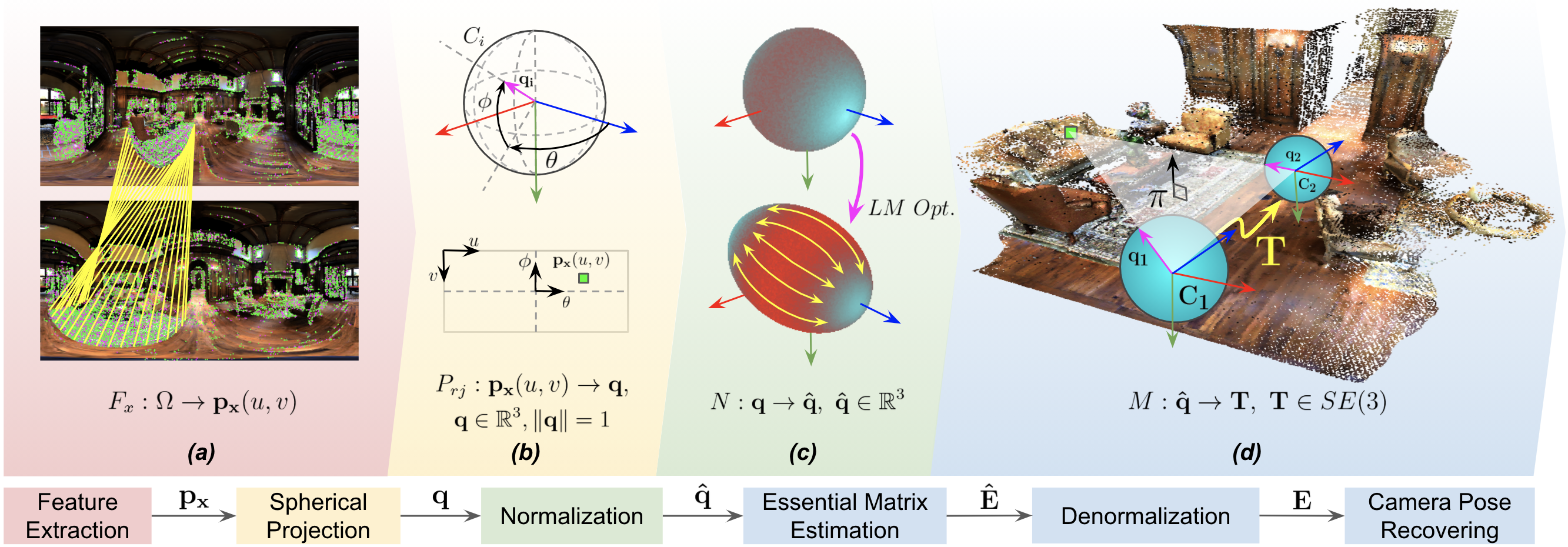}

\captionof{figure}{\textbf{Normalized 8-PA for spherical cameras. }Our method takes two raw 360-FoV images as input, from where we extract several key-features and matching their correspondences across the images as
shown in (a). Next, in (b), the same landmarks are projected into a unit sphere by spherical projection (cf.~\ref{sec:sec_preli}-A). Afterward, the sphere is deformed by our normalization method (cf.~\ref{sec:sec_method}) in order to make a better distribution of key-features (c). 
Lastly, we estimate the essential matrix by using DLT over the normalized domain (cf.~\ref{sec:method:normalization}).
\label{fig:pipeline}}
\end{figure*}

Several works have been proposed to explain the stability of the 8-PA as a linear least-square problem, e.g.,~\cite{wedin1972perturbation, muhlich1998role}. However, most of them assume a known noise distribution in the data. Instead, \cite{da2019perturbation} recently demonstrates that without any noise assumption, the 8-PA stability is highly related to the FoV of the used image. However, the statement assumes a uniform distribution of key-features along the FoV. 

In general, a \kike{key-feature} distribution is highly defined by external factors, such as poor scene illumination, lack of texture, among others~\cite{hartleyBoock2003, book_ma2012invitation, rublee2011orb_features}. Indeed, for spherical cameras, due to the high distortion, uneven feature distributions are a critical issue for matching correspondences as studied in~\cite{taira2015robust_features}. Therefore, using key-features directly from 360-FoV images may define an uneven distribution.
\section{Preliminaries}\label{sec:sec_preli}
\subsection{Spherical Projection and Bearing Vectors}
Unlike perspective images where a homogeneous plane is used, 360-FoV images are generally represented by a centralized spherical projection, which can be described as:

\begin{equation} \label{eq:pxl2sph}
\begin{bmatrix}
\theta \\
\phi
\end{bmatrix}  = \begin{bmatrix}
{2\pi}/{W} && 0 && -\pi \\
0 && {-\pi}/{H} && {\pi}/{2}
\end{bmatrix} \begin{bmatrix}
u \\
v \\
1
\end{bmatrix}
\end{equation}
\begin{equation} \label{eq:sph2bearings}
\mathbf{q_n} = \begin{bmatrix}
x_n \\
y_n \\
z_n
\end{bmatrix} = \begin{bmatrix}
cos(\phi)sin(\theta) \\
-sin(\phi) \\
cos(\phi)cos(\theta) \\
\end{bmatrix},
\end{equation}
where \eqref{eq:pxl2sph} projects a pixel $(u, v)$ into the spherical coordinate $(\theta, \phi)$, and then \eqref{eq:sph2bearings} transforms that coordinate into a unit vector $\mathbf{q_n}$. Hereinafter, we name $\mathbf{q_n}$ as a \textit{bearing vector}. Note that $W$ and $H$ are the width and height of the 360-FoV image.

\subsection{Epipolar Constraint and the Eight-Point Algorithm}
Without loss of generality, the same epipolar constraint defined for perspective images can be applied to spherical projection~\cite{scaramuzza2011tutoI}. Therefore, given a pair of 360-FoV images, two unit sphere cameras can be projected (i.e., $C_1$ and $C_2$ in Fig. \ref{fig:pipeline}-(d)), from which the \textit{bearing vectors} $\mathbf{q_1}$ and $\mathbf{q_2}$ can be also defined. Then, the epipolar constraint, which defines the coplanarity of $\mathbf{q_1}$ and $\mathbf{q_2}$ onto an epipolar plane $\pi$, can be described as follows:
\begin{equation} \label{eq:epipolarConstraint}
\mathbf{q_2^\top E q_1 = 0},\text{ with } \mathbf{E = [t~]_{\times} R}~,
\end{equation}
\kike{where} $\mathbf{E}$ represents the essential matrix $\in \mathbb{R}^{3\times 3}$ with rank of 2; $[\mathbf{t}~]_{\times}$ stands for a skew-symmetric matrix coming from $\mathbf{t} \in \mathbb{R}^3$; and $\mathbf{R}$ defines a relative camera rotation~$\in SO(3)$.

We can then \kike{reformulate} \eqref{eq:epipolarConstraint} into a \textit{Total Least Square Problem} as follows:
\begin{equation} \label{eq:TLS}
    \mathbf{A}[\mathbf{E}]_\textsc{v} = 0~,
\end{equation}
where $[\mathbf{E}]_\textsc{v}$ is a vector of the coefficients in matrix $\mathbf{E}$ by a row-wise concatenation; $\mathbf{A}$ is an $n \times 9$ matrix which is built upon stacking at least $n\geq8$ correspondences of bearing vectors (usually called \kike{\textit{observation matrix}}~\cite{hartleyBoock2003}), where an $i^{th}$ row of this matrix can be defined by the Kronecker product of two correspondence bearing vectors as $\mathbf{A_i} =\mathbf{q^i_1}\otimes \mathbf{q^i_2}$.

Afterward, Singular Value Decomposition (SVD) is applied to $\mathbf{A}$ for finding the solution of \eqref{eq:TLS}, in which the last column of the orthogonal subspace of $\mathbf{A}$ defines $[\mathbf{E}]_\textsc{v}$, and can then be reshaped into $\mathbb{R}^{3\times3}$ and forced into rank(2). This procedure is called \textit{Direct Linear Transformation} (DLT)~\cite{hartleyBoock2003}.
\section{Robust Spherical Normalization} \label{sec:sec_method}

As the key-features are primarily located at areas with high texture, a wider FoV image cannot always provide an uniformly distributed locations of key-features. Thus, we propose our normalization strategy for the classic 8-PA \kike{to reduce the effect} when the key-features are not uniformly distributed.

First, we derive the mechanism introduced in \cite{hartley1997defense} but with our normalization applied, and then we define a new vector space for stabler DLT solution of an essential matrix. We also discuss the reasons why our normalization increases the stability of the 8-PA. Lastly, we present two non-linear optimizations aiming to improve both the 8-PA and GSM methods without overhead computation time.

\vspace{-1mm}
\subsection{Normalization} \label{sec:method:normalization}

\kikereplace{Unlike~\cite{hartley1997defense}, we define our normalization as a non-rigid transformation by a non-zero-determinant matrix, which left-multiplies the bearing vectors and shifts their spatial locations. The normalization is defined as follows:}{
Unlike~\cite{hartley1997defense}, we define our normalization as a transformation that deforms bearing vectors from an unit sphere camera into a ovoid surfaces (see Fig.~ \ref{fig:method}-(c)). This matrix transformation is defined as follows:}
\begin{gather}
    \label{eq:normalization_N}
    \mathbf{N} = \begin{bmatrix}
        S && 0 && 0 \\
        0 && S && 0 \\
        0 && 0 && K 
    \end{bmatrix}~, \\
    \label{eq:bearings_normalization}
    \mathbf{\hat{q}}_i  = \mathbf{N} ~ \mathbf{{q}}_i~,
\end{gather}
where $S, K \in \mathbb{R}$ and $|\mathbf{N}| \neq 0$. For convenience, our normalization is designed as a diagonal matrix~$\mathbb{R}^{3\times 3}$, controlled by two parameters, $S$ and $K$, as presented in~\eqref{eq:normalization_N}, which allows us to deform bearing vectors along XY and Z directions, respectively. Thus, we represent the normalization~\eqref{eq:bearings_normalization} with the epipolar constraint presented in~\eqref{eq:epipolarConstraint}:
\vspace{-1mm}
\begin{gather}
    \label{eq:large_norm_epipolar}
    \mathbf{\hat{q}}_2^\top ~\mathbf{N}_2^{-\top} ~ \mathbf{E} ~ \mathbf{N}_1^{-1} ~ \mathbf{\hat{q}}_1 = 0~.
\end{gather}

By further arranging~\eqref{eq:large_norm_epipolar}, we can build our normalized constraint as follows:
\begin{gather}
    \label{eq:norm_epipolar}
    \mathbf{\hat{q}}_2^\top \hat{\mathbf{E}}~\mathbf{\hat{q}}_1 = 0~, \\
    \label{eq:de_norm_E}
    \mathbf{E} =  \mathbf{N}^{\top}  \hat{\mathbf{E}} ~ \mathbf{N}~.
\end{gather}

Note that~\eqref{eq:norm_epipolar} is embodied by~$\mathbf{\hat{E}}$, which stands for our normalized essential matrix in the normalized domain; thus, we can find~$\mathbf{\hat{E}}$ by using the standard DLT procedure described in Sec.~\ref{sec:sec_preli}. In this way, we can recover the original essential matrix $\mathbf{E}$ by left-right multiplying the matrix $\mathbf{N}$ as described in~\eqref{eq:de_norm_E}, which we call \textit{denormalization}. Finally, the relative camera pose $\mathbf{T} \in SE(3)$ can be recovered from $\mathbf{E}$ by SVD. To be clear, we present the procedure in Fig.~\ref{fig:pipeline}.

\subsection{Deformation of Spherical Projection}
\label{sec:method:def_spherical_domain}

Intuitively, normalizing bearing vectors through a non-rigid transformation~\eqref{eq:normalization_N} deforms the unit sphere into a different spatial domain which in turns defines a new \kike{observation} matrix $\mathbf{A}$. To explain why this normalization leads to more stable DLT solution for~\eqref{eq:norm_epipolar}, we present two properties that define stability for 8-PA in the context of spherical cameras.

\begin{figure*}
\centering
\includegraphics[width=1\linewidth]{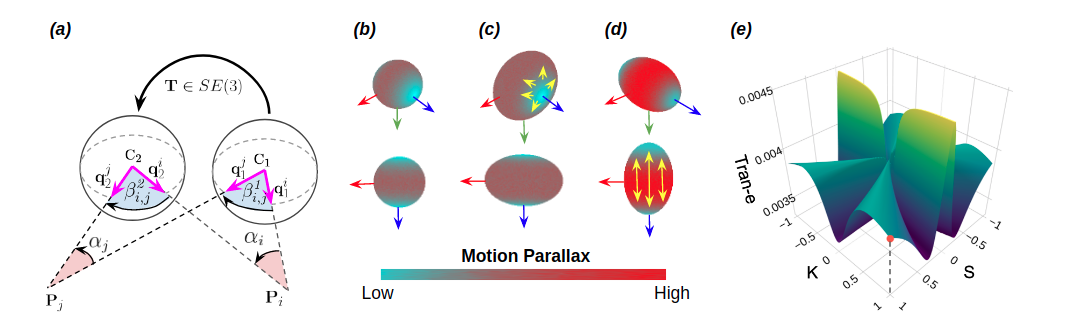}
\caption{
\textbf{Stability of the spherical projection}. In panel (a) given the locations of the landmarks $\textbf{P}_i$ and $\textbf{P}_j$ the internal angles $\beta^1_{ij}$ and $\beta^2_{ij}$ as well as the motion parallax $\alpha_{i}$ and $\alpha_{j}$ are defined.
In panels (b)-(d), geometric visualizations of our normalization are presented (cf. Sec.\ref{sec:method:normalization}). Panel (b) shows a spherical camera without normalization as a reference. In (c), the effects of setting $S=2$ are illustrated, while in (d) the effects of $K=2$. Note that normalizing a spherical camera using $(S,~K)$ can increase the motion parallax in the data.
Lastly, in panel (e) we present a surface loss for translation error build upon several combinations of $S$ and $K$ values from a grid. We can verify that there exists a combination of $(S,~K)$ that reduces error.}
\label{fig:method}
\end{figure*}

First, based on~\cite{da2019perturbation}, we assert that the FoV of an image has a strong correlation with the stability of an essential matrix estimation when using the 8-PA. The main reason is because large FoV images are prone to define large internal angles between its bearing vectors, i.e. the angles $\beta^1_{ij}$ and $\beta^2_{ij}$ in Fig.~\ref{fig:method}(a). This can be mathematically justified by representing these internal angles in terms of the observation matrix $\mathbf{A}$ as shown in \eqref{eq:internal_angles_and_A}, which in turn evaluates a bound for the second least singular value of $\mathbf{A}$ (i.e. $\sigma_8$) as presented in \eqref{eq:sigma_8_and_A}. For more details, we refer to Sec.~3.4. in \cite{da2019perturbation}:
\begin{gather}
    \label{eq:internal_angles_and_A}
    \left \| \mathbf{A}\mathbf{A}^{\top} \right \|^2_F = \sum_i\sum_j{(cos^2_i\beta^1_j)(cos^2_i\beta^2_j)}~, \\
    \label{eq:sigma_8_and_A}
    \sigma_8 \leq \sqrt{ \frac{n}{8} - \frac{1}{8}\sqrt{\frac{8 \left \| \mathbf{A}\mathbf{A}^{\top} \right \|^2_F -n^2 }{7}}}~.
\end{gather}

Based on \cite{da2019perturbation, wedin1972perturbation}, the error in an essential matrix $\mathbf{E}$ can be defined as a function of $\sigma_8$ as follows:
\begin{gather}
    \label{eq:e_bound}
    \Delta \mathbf{E} \leq \frac{\left \| \mathbf{Q} \right \|_2}{\sigma_8}~,
\end{gather}
where $\Delta \mathbf{E}$ is the error in the essential matrix, and $\mathbf{Q}$ represents the perturbation matrix which embodies the noise and outliers in bearing vectors. In practice, based on~\eqref{eq:e_bound}, we assert that larger internal angles, $\beta^1_{ij}$ and $\beta^2_{ij}$, can evaluate larger $\sigma_8$ values, which in turn result in a lower $\Delta\mathbf{E}$ error. Therefore, if we deform the spherical projection by increasing the internal angles between bearing vectors, we will have larger $\sigma_8$ values and thus obtain a more stable solution.
\kike{In addition, based on the translation vector $\mathbf{t}$ between cameras and the distance to some landmarks in the scene,}
we can define the motion parallax for the spherical projection as the angles $\alpha_j$ and $\alpha_i$ in Fig.~\ref{fig:method}(a). Therefore, analogously to the motion parallax defined for perspective projection~(e.g. \cite{hartleyBoock2003, book_ma2012invitation}), we can assert that when the angles $\alpha_j$ and $\alpha_i$ are close to zero, the DLT solution of the 8-PA is unable to recover a camera pose. Thus, the lack of motion parallax is one of the degenerative conditions for the 8-PA~(e.g. \cite{higgins1981computer,hartley1997defense, book_ma2012invitation}). Here, if we properly dislocate every bearing vector in a particular direction, we can modify the motion parallax and further define a more reliable estimation of a camera pose.

As illustrated in Fig.~\ref{fig:method}(c), we compute a grid of $S$ and $K$ values defining different normalizing matrices~\eqref{eq:normalization_N}. By using random synthetic landmarks in 3D, we apply our proposed normalization to two known spherical cameras as described in Sec.~\ref{sec:method:normalization}. Thus, we can visualize that there exists a set of $S$ and $K$ values which improves a camera pose estimation by normalizing bearing vectors.

In the same synthetic environment, we evaluate the motion parallax for every matched bearing vector and plot it in a scale of color in Fig.~\ref{fig:method}(b)-(d). By deforming the spherical projection along a specific direction, we can magnify the motion parallax in a set of bearing vectors.
\vspace{-2mm}
\subsection{Non-linear Optimization over S and K}
\label{sec:method:non_linear_opt}
Although the normalized 8-PA proposed by~\cite{hartley1997defense} gives us a mechanism to apply our normalization~\eqref{eq:normalization_N} to spherical features, the method does not provide a valid procedure to evaluate a normalized matrix under spherical projection.

Therefore, we propose a non-linear optimization to reduce errors in the epipolar constraint by locally finding an optimal matrix $\mathbf{N}$, parameterized by $S$ and $K$, which effectively normalizes bearing vectors in spherical projection.

In practice, based on the projected distance in~\cite{pagani2011_sfm_some_loss_func}, we define our residual error in the epipolar constraint as follows:
\begin{gather}
    \label{eq:projected_residuals}
    \epsilon(\Theta) = \frac{| \mathbf{q}_2^\top~ \mathbf{E}_{\Theta}~\mathbf{q}_1|} {\left \|\mathbf{q}_2 \right \| \left \| \mathbf{E}_{\Theta}~\mathbf{q}_1 \right \|}~,
\end{gather}
where ${\mathbf{E}}_{\Theta}$ represents the essential matrix evaluated by a particular set of $\Theta$ parameters. Thus, for our proposed method, $\Theta$ represents $S$ and $K$ while as for the optimization of GSM, it is defined as $\xi\in \mathbb{R}^{6}$, i.e. $\mathbf{T} \in SE(3)$ mapped into the Lie group $\xi\in\mathfrak{se}(3)$~\cite{book_ma2012invitation}. Thus, we can define our non-linear optimization to find a set of $S^*$ and $K^*$ parameters as follows:
\begin{gather}
    \label{eq:LM_SK}
    S^*,K^* = \argmin_{S, K \in \mathbb{R}^2} \left\| \epsilon (S, K)\right\|^2_1.
\end{gather}

Unlike GSM, which is defined over 6-DoF while minimizing the least-square errors of~\eqref{eq:projected_residuals}, our proposed optimization~\eqref{eq:LM_SK} is defined over 2-DoF only (i.e. $S$ and $K$), which do not evaluate a camera pose directly; thus an initial evaluation of the 8-PA is not needed for an initial guess, which in turns does not add the overhead in time. To deploy our optimization~\eqref{eq:LM_SK}, we leverage the LM algorithm~\cite{levmar} with both $S$ and $K$ parameters set to 1 as a trivial initial point.

\begin{figure*}[t!]
\centering
\includegraphics[width=0.9\linewidth]{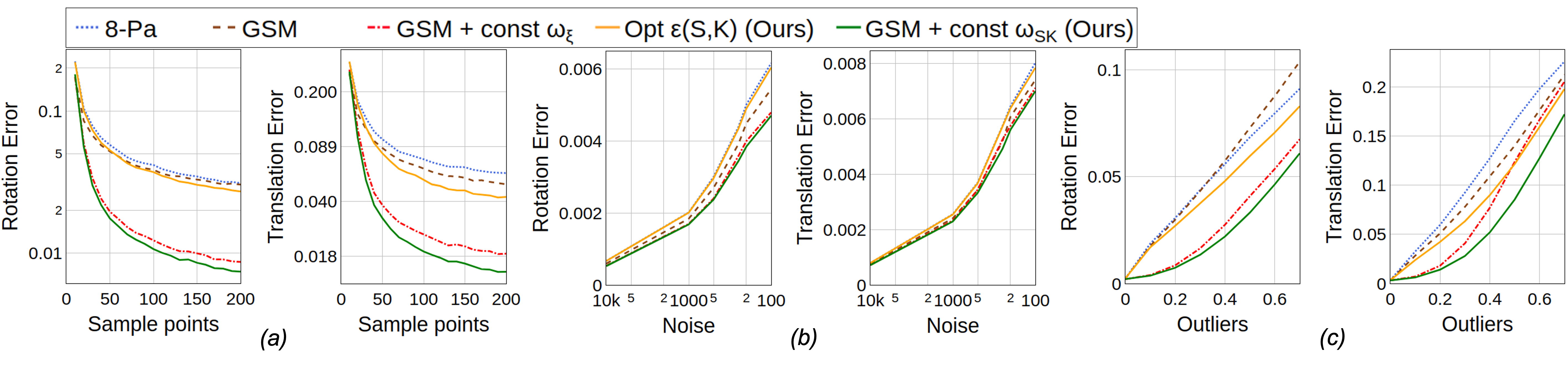}
\caption{
\textbf{Synthetic Points Evaluation.}~ Unless noted otherwise, all of the experiments described in this figure uses: 200 matched/correspondences bearing vectors in spherical projection as input; von Misses-Fisher noise of $\kappa=500$ (3.21$\degree$ of error). In panel (a), a constant outliers ratio of 20\% has been added, while in (b), the input data is evaluated as free-outliers data.
}
\label{fig:outliers}
\end{figure*}

To further show the versatility of our normalization upon the 8-PA solution for 360$^\circ$ images, we also propose a robust constant weighting function to improve the GSM accuracy without increasing its computation time. This optimization is described as follows:
\begin{gather}
    \label{eq:LM_SK_wRT}
    \xi^* = \argmin_{\xi \in \mathbb{R}^6}  \sum_{i} \omega_{i{SK}}~ \epsilon_{i}(\xi)^2~,
\end{gather}
where $\omega_{SK}$ is a constant vector build upon the residuals $\epsilon(S^*, K^*)$, which defines the confidence for every matched bearing vector as a probabilistic function $P(\epsilon(S^*, K^*))$.
Combined with our proposed optimization~\eqref{eq:LM_SK}, we compute a robust essential matrix $\mathbf{E}_{SK}$ and evaluate a residuals vector $\epsilon(S^*, K^*)$, from where $\omega_{SK}$ is computed as a normal distribution $\mathcal{N}(\epsilon(S^*, K^*)|\mu, \sigma)$, evaluating $\mu$ and $\sigma$ as the mean and standard deviation of the residuals vector $\epsilon(S^*, K^*)$, respectively. To deploy the optimization~\eqref{eq:LM_SK_wRT}, the LM algorithm~\cite{levmar} is also used.
\section{Experimental Results} \label{sec:sec_exp}

In this section, we conduct experiments to demonstrate that our method achieves a more accurate camera pose estimation for 360-FoV images. We define two environments: synthetic random points~(Sec.~\ref{sec:exp_noise_outliers}) and sequence of real-images~(Sec.~\ref{sec:exp_ablation} and \ref{sec:exp_seq_images}).

For experiments in Sec.~\ref{sec:exp_seq_images}, we use both 360-FoV images in spherical projection and fish-eye images in the unified camera model~\cite{usenko2018double}. For the former, we collect our own dataset (named 360-MP3D-VO) by rendering the Matterport3D data~\cite{Matterport3D} on Minos~\cite{minos2017} to project a continues sequence of 360-FoV frames; for the latter, we use the TUM-VI dataset~\cite{TUMVIschubert2018dataset}.

The evaluated baselines are the classic 8-PA~\cite{higgins1981computer} in spherical projection and the GSM solution~\cite{pagani2011_sfm_some_loss_func}. For the former, we compare with our normalized approach~\kike{using} the optimization \eqref{eq:LM_SK}, which we refer to as~$Opt~\epsilon(S,K)$. To compare with GSM, we use our proposed optimization~\eqref{eq:LM_SK_wRT}, where we refer to $\omega_\xi$ and $\omega_{SK}$ as a weighting function evaluated by residuals from a camera pose $\xi$ \kike{(computed  by using the 8-PA\cite{hartley1997defense})} and residuals obtained from our normalization $Opt~\epsilon(S,K)$, respectively.

Similar to~\cite{Fathian2018Quest, da2019perturbation, GuanSfM_VMF},
we use $\epsilon_{R}$ as the rotation error and $\epsilon_{t}$ as the translation error for camera pose estimation:
\begin{equation}
\label{eq:metrics}
\epsilon_{R} = \frac{1}{\pi}~cos^{-1}
\left(\frac{tr(\mathbf{R}^\top \mathbf{\Tilde{R}}) - 1}{2}
\right), \quad
\epsilon_\mathbf{t} = \frac{1}{\pi}~ cos^{-1}(\mathbf{t}^\top\mathbf{\Tilde{t}})~.
\end{equation}
\subsection{Noise and Outlier Evaluation}
\label{sec:exp_noise_outliers}

In this experiment, we project several landmarks in 3D by using the ground truth depth data from our 360-MP3D-VO dataset.
To generate ground truth camera poses, we randomly sample translation vectors $\mathbf{t}$ in a uniform range of $[-1, 1]\in\mathbb{R}^3$. Moreover, for each axis, we generate random rotation matrices $\mathbf{R}\in SO(3)$ by sampling Euler-angles in a uniform range of $[-\pi/4, \pi/4]$. Then, we construct our camera poses as homogeneous transformations $\mathbf{T}=[\mathbf{R}|\mathbf{t}]\in SE(3)$.

To show the effect of using different numbers of point, we uniformly sample between 8 to 200 3D-landmarks for each evaluation. Then, similar to~\cite{GuanSfM_VMF, da2019perturbation}, we add a constant von Misses-Fisher noise (vMF) of $\kappa=500$ (i.e., 3.21$\degree$ of error) as well as a constant outlier ratio of 20\%. The results of 15K evaluations are presented in~Fig.~\ref{fig:outliers}-(a).
In addition, to evaluate camera pose under different levels of noise, we sample 200 3D-landmarks of 15K different scenes, and then we incrementally add a vMF noise defined by a $\kappa$ equals to 100, 200, 500, 1000, and 10000, representing 10.21$\degree$, 5.22$\degree$, 3.21$\degree$, 2.27$\degree$, 1.60$\degree$ and 0.72$\degree$ of error, respectively. This experiment is presented in~Fig.~\ref{fig:outliers}-(b).
To further evaluate our method against outliers, we use a constant number of 200 3D-landmarks and a vMF noise of $\kappa=500$, then we increase outliers ratio from 0\% to 70\%. This is shown in~Fig. \ref{fig:outliers}-(c).

In Fig. \ref{fig:outliers}, we show favorable results against others in terms of noise level, outlier level, and number of points. For instance, our normalization is capable of reducing the effect of outliers around 10\% compared with the baseline, i.e., using $\omega_\xi$ without our normalization to build $\omega$.
\vspace{-2mm}
\subsection{Ablation Study}
\label{sec:exp_ablation}

\begin{table}[!t]
\fontsize{18}{20}\selectfont
\caption{\small{Comparisons with 8-PA methods.}}
\label{tab:exp_ablation_1}
\begin{adjustbox}{max width=\columnwidth}
\centering
\begin{tabular}{c|ccccc}
\hline
\hline
    \multicolumn{1}{c}{} &
    \multicolumn{1}{c}{\textbf{$\sigma_8$}$\shortuparrow$} &
    \multicolumn{1}{c}{\textbf{$\alpha$}$\shortuparrow$} & 
    \multicolumn{1}{c}{\textbf{Rot-e $\times 10^{-3}$}} &
    \multicolumn{1}{c}{\textbf{Tran-e $\times 10^{-3}$}} &
    \multicolumn{1}{c}{\textbf{Time (ms)}} \\
\hline
    $Opt~\epsilon(S,K)$ &
    \textbf{20.025} & 0.858 & \textbf{11.429} & \textbf{25.416} & 45.52 \\
    8-PA~\cite{higgins1981computer} & 
    0.650 & 0.731 & 13.387 & 32.367 & \textbf{40.33} \\
    Isotropic~(a)~\cite{hartley1997defense} & 
    1.232 & 0.723 & 15.937 & 49.414 & \textbf{40.33} \\
    Non-Isotropic~(b)~\cite{hartley1997defense} & 
    0.012 & 0.723 & 13.439 & 34.354 & \textbf{40.33} \\
    (a) + (b)~\cite{muhlich1998role} & 
    0.107 & \textbf{0.879} & 14.173 & 38.808 & 40.88 \\
\hline
\hline

\end{tabular}
\end{adjustbox}
\end{table}
\begin{table}[!t]
\fontsize{15}{18}\selectfont
\caption{\small{Ablation study.}}
\label{tab:exp_ablation_2}
\begin{adjustbox}{max width=\columnwidth}
\centering
\begin{tabular}{c|l|ccc}
\hline
\hline
    \multicolumn{1}{c}{} & 
    \multicolumn{1}{c}{} &
    \multicolumn{1}{c}{\textbf{Rot-e $\times 10^{-3}$}} & 
    \multicolumn{1}{c}{\textbf{Tran-e $\times 10^{-3}$}} & 
    \multicolumn{1}{c}{\textbf{Time (ms)}} \\
\hline
    \multicolumn{2}{c|}{GSM~\cite{pagani2011_sfm_some_loss_func} (a)} & 9.1029 & 19.7902 & 45.318 \\
\hline
    \multirow{4}{*}{Gaussian}
    & (a) + not const~$\omega_\xi$ & 2.0951  & 5.2065  & 218.598  \\
    & (a) + not const~$\omega_{SK}$ & \textbf{2.0943}  & \textbf{5.1894}  & 212.924 \\
    & (a) + $\omega_\xi$ & 3.9048  & 9.2592  & \textbf{50.279}  \\
    & (a) + $\omega_{SK}$ & 3.6802  & 7.5717  & 58.651  \\
\hline
    \multirow{4}{*}{t-distribution} 
    & (a) + not const~$\omega_\xi$ & 9.1810  & 19.8160 & \textbf{48.126}  \\
    & (a) + not const~$\omega_{SK}$ & \textbf{8.9496} & \textbf{18.9284}  & 58.836  \\
    & (a) + $\omega_\xi$ & 9.1810 & 19.8160  & 50.279  \\
    & (a) + $\omega_{SK}$ & 9.0581 & 19.5573  & 53.795 \\
\hline
\hline
\end{tabular}
\end{adjustbox}
\end{table}

\begin{table*}[t!]
\vspace{2mm}
\fontsize{8}{11}\selectfont
\caption{\small{Experimental results in real scenes on MP3D-VO and TUM-VI.}}
\begin{adjustbox}{max width=\textwidth}
\centering
\begin{tabular}{ c | c | cc | cccccc | ccc}

\hline
\hline

\multicolumn{2}{c}{} &
\multicolumn{5}{c}{\textbf{Rotation error $\times 10^{-3}$}} &
\multicolumn{1}{c}{} &
\multicolumn{5}{c}{\textbf{Translation error $\times 10^{-3}$}} \\
\cline{3-7} \cline{9-13}

\multicolumn{1}{c}{} &
\multicolumn{1}{c}{} &
\multicolumn{1}{c}{8-PA~\cite{higgins1981computer}} &
\multicolumn{1}{c}{$Opt~\epsilon(S, K)$} &
\multicolumn{1}{c}{GSM~\cite{pagani2011_sfm_some_loss_func}} &
\multicolumn{1}{c}{$\omega_{\xi}$} &
\multicolumn{1}{c}{$\omega_{SK}$} &
\multicolumn{1}{c}{} &
\multicolumn{1}{c}{8-PA~\cite{higgins1981computer}} &
\multicolumn{1}{c}{$Opt~\epsilon(S, K)$} &
\multicolumn{1}{c}{GSM~\cite{pagani2011_sfm_some_loss_func}} &
\multicolumn{1}{c}{$\omega_{\xi}$} &
\multicolumn{1}{c}{$\omega_{SK}$} \\
\hline




\multirow{3}{*}{MP3D-VO} 
& Q75 & 
23.577 & \textbf{20.504} & 15.286 & 9.909 & \textbf{7.964} &  & 91.442 & \textbf{72.693} & 43.707 & 29.945 & \textbf{20.653} \\

& Q50 & 
10.827 & \textbf{9.515}  & 7.389  & 3.523 & \textbf{3.072} &  & 36.814 & \textbf{29.645} & 19.453 & 9.985  & \textbf{7.966}  \\

& Q25 & 
4.624  & \textbf{4.075}  & 3.277  & 1.638 & \textbf{1.524} &  & 14.745 & \textbf{12.276} & 8.501  & 4.611  & \textbf{4.009} \\
\hline

\multirow{3}{*}{TUM-VI \cite{TUMVIschubert2018dataset}} 
& Q75 &
44.479 & \textbf{40.514} & 49.291 & 39.938 & \textbf{38.256} &  & 221.131 & \textbf{199.042} & 166.594 & 169.122 & \textbf{140.846} \\

& Q50 &
27.211 & \textbf{24.969} & 27.226 & 21.075 & \textbf{19.241} &  & 117.701 & \textbf{104.742} & 82.457  & 74.661  & \textbf{62.743}  \\

& Q25 &
15.512 & \textbf{14.462} & 14.595 & 10.507 & \textbf{9.708}  &  & 58.791  & \textbf{53.283}  & 40.899  & 33.724  & \textbf{28.800}  \\

\hline
\hline

\end{tabular}
\end{adjustbox}
\label{tab:exp_scenes}
\vspace{-2mm}
\end{table*}
\begin{table*}[!ht]
\fontsize{10}{13}\selectfont
\caption{\small{Experimental results under different thresholds with RANSAC.}}

\begin{adjustbox}{max width=\textwidth}
\centering
\begin{tabular}{ c | cc | cccccc | ccc }

\hline
\hline

\multicolumn{1}{c}{} &
\multicolumn{5}{c}{\textbf{Threshold:} 2.30E-04 \  \textbf{Iterations:} 590 \ \textbf{Outliers in data:} 1\%} &
\multicolumn{1}{c}{} &
\multicolumn{5}{c}{\textbf{Threshold:} 1.10E-03 \ \textbf{Iterations:} 66 \ \textbf{Outliers in data:} 20\%} \\
\cline{2-6}
\cline{8-12}

\multicolumn{1}{c}{} &
\multicolumn{1}{c}{8-PA~\cite{higgins1981computer}} &
\multicolumn{1}{c}{$Opt~\epsilon(S, K)$} &
\multicolumn{1}{c}{GSM~\cite{pagani2011_sfm_some_loss_func}} &
\multicolumn{1}{c}{$\omega_{\xi}$} &
\multicolumn{1}{c}{$\omega_{SK}$} &
\multicolumn{1}{c}{} &
\multicolumn{1}{c}{8-PA~\cite{higgins1981computer}} &
\multicolumn{1}{c}{$Opt~\epsilon(S, K)$} &
\multicolumn{1}{c}{GSM~\cite{pagani2011_sfm_some_loss_func}} &
\multicolumn{1}{c}{$\omega_{\xi}$} &
\multicolumn{1}{c}{$\omega_{SK}$} \\
\hline

\multirow{1}{*}{Rot-e $\times 10^{-3}$} 
& 3.957 & \textbf{3.950} & 3.691 & 2.994 & \textbf{2.919} &  & 10.658 & \textbf{9.963} & 9.978 & 4.292 & \textbf{4.065} \\

\multirow{1}{*}{Tran-e $\times 10^{-3}$} 
& 8.757 & \textbf{8.689} & 8.239 & 7.660 & \textbf{7.477} &  & 22.656 & \textbf{19.806} & 20.394 & 10.540 & \textbf{9.452} \\
\hline

\multirow{1}{*}{Residual} 
& \textbf{2.72E-06} & \textbf{2.72E-06} & \textbf{2.88E-06} & 2.94E-06 & 2.94E-06 &  & \textbf{2.72E-05} & 2.74E-05 & \textbf{2.77E-05} & 2.91E-05 & 2.92E-05 \\
\hline


\multirow{1}{*}{Time (sec)} 
& \textbf{0.149} & 0.153 & \textbf{0.181} & 0.183 & 0.189 &  & \textbf{0.015} & 0.021 & \textbf{0.050} & 0.065 & 0.055 \\
\hline
\hline


\end{tabular}
\end{adjustbox}
\vspace{-5mm}
\label{tab:exp_threshold}
\end{table*}
\vspace{-1mm}

In this experiment, we show results of our solution by using tracked key-features from a sequence of 360-FoV images in a 6-DoF camera motion. For each frame in this sequence, we extract around 200 key-features using Shi-Tomasi key-points~\cite{tommasini1998making} and the KLT tracker~\cite{lucas1981iterative}. For every evaluation, we ensure that there exits at least a minimum distance of 0.5m between frames. We evaluate over 2K different environments and compute the median values of the estimated errors.

In Table~\ref{tab:exp_ablation_1}, we compare the proposed normalized solution with the 8-PA algorithms~\cite{higgins1981computer, hartley1997defense, muhlich1998role}, all of them in spherical projection. Note that isotropic and non-isotropic pre-conditioning are the normalization strategies proposed by~\cite{hartley1997defense}, which are particularly used for uncalibrated cameras in perspective view, but can still be used for spherical projection. In the results, our normalization is capable of reducing errors in camera pose without significantly adding the overhead time. Note that, our normalization obtains larger values of $\sigma_8$ (second least singular value of an observation matrix $\mathbf{A}$) as well as $\alpha$ (motion parallax in the normalizing domain), showing that our normalization truly increases the stability in the DLT solution of an essential matrix.

In Table~\ref{tab:exp_ablation_2}, we compare the effect of our normalization in the weighted non-linear optimization~\eqref{eq:LM_SK_wRT} upon the GSM solution~\cite{pagani2011_sfm_some_loss_func}.
In rows 2-5, we find that using a Gaussian distribution constantly performs better than t-distributions. In rows 2,3 and 6,7, we evaluate a non-constant weighing function $\omega$ which is updated at every iteration in the LM optimization~\cite{torr1997developmentIRLS, kerl2013robust, GuanSfM_VMF}; this approach is also known as Iterative Re-weighted Least-Square method (IRLS). Although IRLS achieves the lowest error, it is still the most time-consuming method among all evaluations.

\vspace{-2mm}
\subsection{Experiments on Real Scenes}
\label{sec:exp_seq_images}
We evaluate a sequence of 360-FoV and fish-eye images with our own 360-MP3D-VO and the TUM-VI~\cite{TUMVIschubert2018dataset} datasets, respectively. Similar to Sec.~\ref{sec:exp_ablation}, we track 200 key-features between frames, evaluating around 15K pairs of frames using 360-MP3D-VO, and 16K paired images using TUM-VI.

For the evaluations on the TUM-VI~\cite{TUMVIschubert2018dataset} dataset, we only use the scenes that contain a complete ground truth camera pose, i.e., room-1 to room-6. Moreover, since this dataset is mainly used for visual-inertial tasks, some frames in our evaluation has been skipped due to drastic camera movements and severe changes in illumination.

Results of camera pose errors on both datasets are presented in Table \ref{tab:exp_scenes}, using the quantiles Q25, Q50 and  Q75 of error evaluations.
From the results, we can verify that our approaches constantly outperform the baselines, with the lowest errors in every evaluation. Moreover, for the averaged condition (i.e., Q50 results), our strategies are capable of reducing errors up to 10\% by normalizing bearing vectors, using $Opt~\epsilon(S, K)$; and up to 50\% by using our weighted optimization $\omega_{SK}$.

\kike{Additionally,} we evaluate our proposed approaches $Opt~\epsilon(S,K)$ and $\omega_{SK}$ in the context of a RANSAC evaluation. We design two settings using the same amount of correspondence bearing vectors, noise, and outliers (e.g., 400 3D-landmark, vMF noise of $\kappa=500$, and 50\% of outliers), but with two different thresholds for RANSAC. In addition, upon the RANSAC results, we evaluate the final essential matrix using only the detected inliers by using our proposed methods as well as the baselines. In Table~\ref{tab:exp_threshold}, results over 2K evaluations are presented.

On the left side of Table~\ref{tab:exp_threshold}, we set a threshold for RANSAC as $2.3\times10^{-4}$, which successfully detects all the inliers after 590 iterations; whereas on the right side, the previous threshold is relaxed until $1.1\times10^{-3}$, speeding up the evaluation by detecting 70\% of inliers in 66 RANSAC iterations. However, since we know in advance the outlier ratio of our data, we can assert that there is 20\% of the remaining outliers. Therefore, comparing columns 2 and 11, we verify that our proposed $\omega_{SK}$ performs similar to the one using RANSAC with a tight threshold, but 3 times faster, showing the benefit of our robust estimation.

\vspace{-1mm}
\section{Conclusions}\label{sec:sec_concl}
\vspace{-1mm}

In this paper, we propose a novel pre-conditioning strategy to the classic 8-point algorithm~\cite{higgins1981computer} for estimating an essential matrix in spherical projection.
Our solution redesigns the normalizing algorithm~\cite{hartley1997defense} to alleviate the effect of poor/uneven distribution of key-features, increasing stability and robustness against outliers.
We also extend our approach, towards improving the well-known Gold Standard Method~\cite{pagani2011_sfm_some_loss_func} for spherical projection by proposing a constant weighted non-linear optimization, built upon our normalization strategy. Based on extensive experiments under different scene conditions, our proposed methods outperform the baselines, increasing the accuracy in camera pose up to 30\% without a significant impact on the computation time.
\vspace{-2mm}
{\flushleft {\bf Acknowledgement.}} This project is supported by MOST Joint Research Center for AI Technology and All Vista Healthcare with program MOST 110-2634-F-007-016 and MOST 110-2636-E-009-001.

{\small
\bibliographystyle{ieee_fullname}

}

\end{document}